\documentclass{article} 
\usepackage[final]{colm2026_conference}

\usepackage{microtype}
\usepackage{hyperref}
\usepackage{url}
\usepackage{booktabs}
\usepackage{graphicx}
\usepackage{amsmath}
\usepackage{amssymb}
\usepackage{enumitem}
\usepackage{multirow}
\usepackage{array}
\usepackage{tcolorbox}
\newtcolorbox{promptbox}{colback=gray!5, colframe=gray!50, fontupper=\small\ttfamily, left=4pt, right=4pt, top=4pt, bottom=4pt}

\usepackage{lineno}

\definecolor{darkblue}{rgb}{0, 0, 0.5}
\hypersetup{colorlinks=true, citecolor=darkblue, linkcolor=darkblue, urlcolor=darkblue}

\title{PromptNCE: Conditional Probabilities and PMI \\
Using Only LLMs and Contrastive Estimation Prompts}

\author{
Juliette Woodrow \\
Department of Computer Science \\
Stanford University \\
\texttt{jwoodrow@cs.stanford.edu}
\And
Chris Piech \\
Department of Computer Science \\
Stanford University \\
\texttt{piech@cs.stanford.edu}
}

\definecolor{newblue}{rgb}{0.0,0.30,0.95}
\definecolor{cutorange}{rgb}{0.85,0.33,0.0}

\begin{document}

\ifcolmsubmission
\linenumbers
\fi

\maketitle

\begin{abstract}

Estimating mutual information from text usually requires training a task-specific critic, which limits its use in low-data settings. We ask whether large language models can instead estimate pointwise mutual information zero-shot, using only prompts and elicited probabilities. We construct a benchmark from three publicly available human-annotated datasets with ground-truth PMI, and evaluate five information-theoretic prompting-based estimators. Our main method, \textsc{PromptNCE}, frames conditional probability estimation as a contrastive task and augments the candidate set with an explicit OTHER category. The OTHER category allows the model to assign probability mass outside the candidate set, avoiding the closed-set normalization of standard contrastive prompts. \textsc{PromptNCE} gives the best conditional probability estimates on all three datasets. For full PMI, we find that estimating label base rates is the primary bottleneck on two of the three datasets, with the best methods reaching Spearman correlation up to 0.78. We also present a case study in computer science education showing how these estimators can be used to score student knowledge summaries in a low-data setting. We release our code and prompts.

\end{abstract}

\section{Introduction}

Mutual information (MI) between two random variables $X$ and $Y$ quantifies how much knowing one reduces uncertainty about the other. Its pointwise variant, pointwise mutual information (PMI), measures this for a specific pair $(x, y)$ rather than in expectation, making it the natural quantity for scoring individual text pairs by how informative they are about each other. MI and PMI appear throughout machine learning, from learning representations \citep{hjelm2019learning} to feature selection \citep{vergara2014review} to analyzing what neural networks learn \citep{shwartz2017opening}. We consider the concrete case of scoring how much one piece of text tells you
about another when there is no ground truth to score against. We use PMI as an
evaluation signal rather than a training objective. Table~\ref{tab:benchmark}
makes this setting concrete and shows the benchmark datasets we use to evaluate
it with human-derived ground truth. Estimating these quantities is notoriously difficult. The relevant distributions are often high-dimensional and long-tailed, and exact computation is intractable for all but the simplest cases. Decades of work have addressed this by training neural critics to optimize variational bounds on MI \citep{belghazi2018mine, nguyen2010estimating, oord2018representation}. These methods are effective but require extensive task-specific training data, which limits their use in low-data settings. 

Large language models offer a potential path forward. Recent work shows they can serve as effective zero-shot scorers and judges on tasks well-covered by their training data \citep{liu2023geval, zheng2023judging, li2024generation, li2024llmsjudges}. In this paper, we study whether large language models can estimate pointwise mutual information using only prompts and elicited probability estimates. We consider two strategies. The first is to simply ask the model to estimate PMI for a given pair. The second is theory-driven where we translate classical mutual information estimation ideas into natural language prompts, treating the language model as a critic that elicits the probabilities needed to estimate PMI. We formalize this as the \emph{Zero-Shot LLM Mutual Information Estimation Challenge} and construct a benchmark from three publicly available datasets with human-derived ground-truth PMI. We develop five prompting-based approaches for estimating PMI with language models, including direct estimation, decomposition into conditional and marginal terms, and contrastive estimation. Our main method, PROMPTNCE, augments the contrastive candidate set with an explicit OTHER category. Standard contrastive approaches inflate conditional estimates. The \textsc{other} category lets the model express that most mass falls outside the candidates, and this is designed to better approximate the open-vocabulary conditional. \textsc{PromptNCE} achieves the best conditional estimates on all three datasets, though on full PMI the decomposed methods largely converge. We find that performance depends strongly on dataset structure. We introduce a variance decomposition that separates how much PMI variation comes from each term and show that it predicts when methods succeed and when they fail (Section \ref{sec:datasets}).  

\begin{table}[t]
\centering
\footnotesize
\setlength{\tabcolsep}{3pt}
\begin{tabular}{@{}l >{\raggedright\arraybackslash}p{4.15cm} ccc ccc@{}}
\toprule
& \multicolumn{4}{c}{One example pair $x \rightarrow y$} & \multicolumn{3}{c}{Dataset structure} \\
\cmidrule(lr){2-5}\cmidrule(l){6-8}
Dataset & Example & $P(y \mid x)$ & $P(y)$ & PMI & $|\mathcal{Y}|$ & $R$ & $\rho_{\mathrm{marg}}$ \\
\midrule
Words
& \texttt{SHOVE} $\rightarrow$ \texttt{PUSH}
& 0.94 & 0.0005 & 7.54
& 10{,}469 & 3.86 & 0.86 \\[3pt]

ChaosNLI
& ``a boy walks through the water.'' / ``A boy walks on water.''
  $\rightarrow$ contradiction
& 0.92 & 0.199 & 1.53
& 3 & 0.07 & $-$0.03 \\[3pt]

GoEmotions
& ``That game hurt.'' $\rightarrow$ sadness
& 0.40 & 0.033 & 2.51
& 28 & 2.69 & 0.77 \\
\midrule
\multicolumn{8}{@{}p{12.4cm}@{}}{\textit{Case study} (\S\ref{sec:case_study}):
$x$ is a paragraph summarizing one student, $y$ is a program she later wrote.
$\mathcal{Y}$ is unbounded and each pair is observed once, so there is no
ground-truth PMI to count.} \\
\bottomrule
\end{tabular}

\caption{One example pair per dataset with its human-derived ground truth,
alongside the structure of its dataset. $R = \mathrm{Var}[\log P(y)] /
\mathrm{Var}[\log P(y \mid x)]$. $R > 1$ means PMI variation is carried by the
marginal. $\rho_{\mathrm{marg}}$ is Spearman $\rho$ from ranking pairs on the
empirical $\log P(y)$. PMI in nats.}
\label{tab:benchmark}
\end{table}

Finally, we present a case study that illustrates why zero-shot PMI estimation matters in practice. LLMs can generate natural-language summaries about individuals from observed data, such as a summary of what a student understands based on the code she has written, or a summary of a patient based on her medical records. But evaluating the quality of such a summary remains an open problem. A useful summary should contain enough individual-specific information to help predict what that person will do next, exactly the kind of relationship PMI captures. In section \ref{sec:case_study}, we show that \textsc{PromptNCE} can provide a principled scoring signal for summaries in one low-data setting when traditional PMI estimation methods would be infeasible. We make the following \textbf{contributions}:
\begin{itemize}[leftmargin=2em]

    \item \textbf{Problem and benchmark.} We formalize the Zero-Shot LLM Mutual Information Estimation Challenge and construct a benchmark from existing human-annotated datasets. 
    
    \item \textbf{PromptNCE.} We introduce \textsc{PromptNCE}, a contrastive prompting method that adds an explicit \textsc{other} category to the candidate set. We show that adding a residual category allows the model to assign probability outside the closed candidate set. 

    \item \textbf{Empirical results.} We show that \textsc{PromptNCE} gives the best conditional estimate on all three datasets, and that base-rate estimation is the bottleneck for full PMI on marginal-dominated datasets.

    \item \textbf{Error analysis.} We introduce a variance decomposition that separates how much of a dataset's PMI variation comes from each term, and show that it predicts when marginal estimation matters. We further show that ranking errors are systematic rather than stochastic, pointing to limits in the model's distributional knowledge.

    \item \textbf{Case study.} We present a case study in CS education showing that
\textsc{PromptNCE} provides a principled evaluation for natural-language summaries
of student understanding and reliably scores expert summaries above generic ones.

    \item \textbf{Code and prompts.} We release our code and prompts at \url{https://juliettewoodrow.github.io/promptnce/}.
\end{itemize}

\subsection{The Zero-Shot LLM Mutual Information Estimation Challenge}
\label{sec:challenge}

The \textit{Zero-Shot LLM Mutual Information Estimation Challenge} is to estimate pointwise mutual information for text pairs using only a pretrained language model and a prompt. The model receives a natural language description of the task and returns a verbal estimate. Consider the word-association pair \texttt{SHOVE} $\rightarrow$ \texttt{PUSH} from Table~1. Estimating its PMI requires knowing both how likely a person is to respond \texttt{PUSH} when prompted with \texttt{SHOVE}, $P(\texttt{PUSH} \mid \texttt{SHOVE})$, and how often \texttt{PUSH} is produced across all cues, $P(\texttt{PUSH})$. Neither quantity is directly available to a language model. Unlike traditional MI estimators, it has no task-specific critic trained to estimate these probabilities from the target distribution. Instead, we ask whether they can be recovered from the model's distributional knowledge through prompting. The marginal poses an additional challenge because it is specific to the target dataset. A model may have a reasonable sense of how common \texttt{PUSH} is in general, but the relevant quantity is how frequently \texttt{PUSH} appears as a response in this particular word-association dataset. The model has no direct access to these frequencies, so its prior knowledge may not match the distribution whose PMI we want to estimate. To evaluate pointwise mutual information for text pairs, we use Spearman rank correlation between estimated and ground-truth PMI, since the most natural downstream use of PMI is identifying which label $y$ has the highest mutual information with a given input $x$. A ``good'' zero-shot MI estimator should rank pairs correctly by PMI.

\subsection{Background and Related Work}

The standard approach to MI estimation is to optimize a variational lower bound using a learned critic. The Donsker-Varadhan (DV) representation gives a tight bound but requires estimating an intractable log-partition function. \citet{belghazi2018mine} operationalized this idea in MINE by training a neural critic with gradient descent. \citet{nguyen2010estimating} introduced the NWJ bound, which gives a looser but often lower-variance objective. Contrastive methods offer a simpler alternative. \citet{oord2018representation} showed that identifying the true pair among distractors gives a lower bound on MI, an objective widely adopted in representation learning \citep{gutmann2010noise,hjelm2019learning,chen2020simple}. One key limitation shared by these methods is that the critic function needs extensive training data and specific modeling assumptions. We instead ask whether an LLM can serve as the critic without any training, by prompting it to provide the probability estimates needed for PMI directly.

Prompting is not the only way to estimate mutual information without a critic. \citet{xu2025gem} introduce GEM, a generative estimator that scores a candidate response by how much it raises the model's likelihood of a reference response, giving a mutual-information estimate that needs no gold standard. GEM needs no learned critic, but it does require white-box access to per-token probabilities. It is the closest prior work to ours and we adopt it as a baseline. Our contribution differs in mechanism: we elicit probabilities through a prompt, which needs only black-box access and lets the estimated quantity be specified in natural language, whereas GEM reads them off the decoder.

Whether LLMs can produce usable probability estimates from prompts alone remains an open question. \citet{tian2023just} found that for RLHF models, verbalized confidence scores are substantially better calibrated than token-level probabilities. \citet{xiong2024can} evaluated black-box confidence elicitation across LLMs, finding persistent overconfidence that no single prompting technique consistently resolves. \citet{chen2024quantifying}, \citet{wang2024calibrating}, and \citet{kapoor2024must} have since explored sampling-based and training-based approaches for more reliable uncertainty expression. Prior work treats verbalized probabilities as confidence in a final answer. We instead use them as plug-in estimates of the terms needed to compute PMI, so the relevant failure modes are in the model's distributional knowledge rather than its calibration.

Whether that knowledge is sufficient for zero-shot estimation is an empirical question, but recent evidence is encouraging. A growing literature uses LLMs as zero-shot scorers  \citep{liu2023geval,zheng2023judging,li2024generation,li2024llmsjudges}. LLM judges can match human agreement rates above 80\% on tasks well-represented in pretraining, though they remain susceptible to bias and degraded performance on specialized criteria \citep{zheng2023judging,li2024llmsjudges}. Our setting differs in that we do not ask for a direct quality judgment. Instead, we take classical MI estimation methods and recast them as prompts, using elicited probabilities as plug-in estimates for the terms in PMI. To the best of our knowledge, this is the first work to use prompted probability elicitation for pointwise mutual information estimation.

\section{Theoretical Motivation}
\label{sec:theory}

Our goal is to estimate pointwise mutual information,
\begin{align}
\mathrm{PMI}(x,y) = \log P(y \mid x) - \log P(y),
\label{eq:pmi}
\end{align}

using only zero-shot probability elicitation from a language model. This 
section develops the theoretical structure underlying our methods. We begin 
by viewing PMI as a difference of two terms that can, in principle, be 
elicited independently. We then show that a contrastive formulation---asking  the model to identify a true label among distractors---recovers PMI exactly up to a candidate-set-dependent constant. Finally, we motivate augmenting the candidate set with an explicit OTHER category to account for probability mass outside the candidates to better estimate the open-vocabulary conditional $P(y \mid x)$.

\subsection{PMI as a Probability Decomposition}
\label{sec:pmi_decomposition}

The identity
\[
\operatorname{PMI}(x,y) = \log P(y \mid x) - \log P(y)
\]
reduces PMI estimation to two quantities: the conditional $\hat{P}(y \mid x)$ and marginal $\hat{P}(y)$. This decomposition is useful because it separates two distinct estimation tasks. The conditional term $P(y \mid x)$ captures how the input $x$ shifts probability across outputs, while the marginal term $P(y)$ captures how common $y$ is overall. Returning to \texttt{SHOVE} $\rightarrow$ \texttt{PUSH}, the conditional asks how often \texttt{PUSH} follows \texttt{SHOVE}, while the marginal asks how often \texttt{PUSH} occurs as a response across all cue words. Errors from either component propagate into the final PMI estimate, and their relative importance depends on the problem setting. In some settings, most of the variation in PMI across pairs comes from the marginal term, while in others the conditional term dominates. We quantify this empirically for each dataset via a variance decomposition in Section~4.1. 

\textbf{Estimating the marginal:} The marginal is a property of the target dataset, but the model has no direct access to its label frequencies. Asking directly for $P(y)$ therefore requires the model to infer how common $y$ is from the task description and its general distributional knowledge, which may not match the target distribution. For example, the model's general knowledge of how common the word \texttt{PUSH} is need not reflect how often people produce \texttt{PUSH} in a word-association experiment. To better ground this estimate, our marginal prompt provides four input-output pairs from the target dataset, along with label names sampled uniformly from the label space (eight for Words and GoEmotions, three for ChaosNLI). These examples do not reveal label frequencies or PMI values, but provide information about the domain, label space, and style of the target distribution, making the marginal estimate less dependent on generic prior knowledge and better aligned with the target dataset. Because this grounding draws on the target dataset, \textsc{PromptNCE} is zero-training and low-data rather than strictly zero-shot. Only the marginal term uses these examples. In our main comparison, every decomposed method uses the same grounded marginal prompt, ensuring equal access to this information.

\subsection{PMI from Contrastive Estimation}
\label{sec:contrastive}

Eliciting a well-calibrated value of $P(y \mid x)$ directly from a language 
model is difficult: the model must assign an absolute probability to a single 
label without any reference point. A contrastive formulation offers a more 
tractable path. Rather than requesting an absolute probability, we ask the 
model to identify the true label among $K-1$ distractors. This is a comparative
judgment grounded in a concrete set of alternatives. This reformulation might 
seem to discard the probabilistic structure we need, but the opposite is true: 
as we now show, the posterior over this identification task recovers pointwise 
mutual information exactly, up to a constant that depends only on the candidate 
set.

Construct a candidate set $S = \{y_1, \dots, y_K\}$, formed by placing the 
true label $y$ at a uniformly random index $I = i^*$ and filling the remaining 
$K-1$ slots with distractors drawn from $P(y)$. We task the model with 
recovering $i^*$ from $S$ given $x$.

To understand what this task reveals, we restate the derivation of 
\citet{oord2018representation} for the posterior over $I$ under this 
generative model.
The joint probability of a configuration in 
which $i$ is the true index is proportional to 
$P(x, y_i) \prod_{j \neq i} P(y_j)$.
Forming the posterior, the factor $\prod_j P(y_j)$ is common to all 
terms and cancels, as does $P(x)$, yielding
$P(I = i \mid x, S) \propto P(x, y_i)/P(y_i) \propto P(y_i \mid x)/P(y_i)$.
Normalizing over $i$ gives
\begin{align}
  P(I = i \mid x, S)
  \;=\;
  \frac{P(y_i \mid x) / P(y_i)}
       {\sum_{j=1}^{K} P(y_j \mid x) / P(y_j)}.
\label{eq:nce}
\end{align}
We now use equation~\eqref{eq:nce} to recover pointwise mutual 
information directly. Since $P(y \mid x)/P(y) = \exp(\mathrm{PMI}(x,y))$ 
by definition, 
equation~\eqref{eq:nce} is a softmax over PMI scores. Evaluating at 
the true label $y$ at index $i^*$ and inverting gives
\[
\mathrm{PMI}(x, y) = \log P(I = i^* \mid x, S) + \log Z(x, S),
\]
where $Z(x,S) = \sum_{j=1}^K \exp(\mathrm{PMI}(x,y_j))$ depends only 
on $S$.
Thus, the contrastive posterior recovers $\mathrm{PMI}(x, y)$ up to an 
additive constant shared across all candidates in $S$. This is sufficient for 
relative comparisons, and in particular for ranking candidates by PMI. However, this estimate will be biased by any changes to $S$, including changing the target $y$. 

\subsection{Recovering Conditional Probabilities from Contrastive Estimation}
\label{sec:residual}

We now use contrastive estimation to estimate the open-vocabulary conditional $P(y \mid x)$, enabling PMI comparisons across different candidate sets $S$.

One might hope to use $P(I = i^* \mid x, S)$ directly as an estimate of
$P(y \mid x)$. This is tempting because from equation~\eqref{eq:nce}, the
posterior is proportional to $P(y \mid x) / P(y)$: if $P(y)$ were roughly constant across candidates, renormalization would approximately recover the conditional. But marginals vary substantially across candidates. Moreover, asking a model to distribute probability over a finite set $S$ forces all mass onto the listed candidates regardless of how much weight the true conditional places outside $S$. Even under this approximation, the result is not $P(y \mid x)$ but its renormalization onto $S$:
\[
P_S(y \mid x) \;=\; \frac{P(y \mid x)}{\sum_{y' \in S} P(y' \mid x)}.
\]
Since the denominator is strictly less than one whenever $P(Y \notin S \mid x)
> 0$, this systematically overestimates $P(y \mid x)$.

We address this closed-set normalization by introducing a residual category \textsc{other} $= \mathcal{Y} \setminus S$ and asking the model to distribute probability over $S \cup \{\textsc{other}\}$. The \textsc{other} category provides an explicit destination for the residual mass $P(Y \notin S \mid x)$, so the model is not forced to renormalize the conditional onto $S$.

In practice, this requires only a single extra category in the prompt, giving
\[
\hat{P}(y \mid x)
\approx
\hat{P}(I = i^* \mid x, S \cup \{\textsc{other}\}).
\]
This conditional estimate is combined with a marginal term to predict PMI as
shown in Equation~\ref{eq:pmi}.

\section{Methods for Estimating Pointwise Mutual Information}
\label{sec:methods}

Section~\ref{sec:theory} motivates the quantities our prompting methods aim to estimate. In practice, we elicit estimates of these quantities by prompting an LLM. We present five prompting methods for estimating PMI, which differ in how they elicit the quantities needed for PMI from the model. 
The experiments that follow evaluate how well each method estimates ground-truth PMI and its components. We organize the methods from simplest to most structured. For a controlled comparison of conditional elicitation strategies, all decomposed methods that estimate a marginal---Direct-Split, MarginalNCE, and \textsc{PromptNCE}---use the same grounded marginal prompt described in Section~\ref{sec:pmi_decomposition}. Thus, these methods differ only in how they elicit the conditional term. Prompts are included in our code release, and Appendix~\ref{app:prompts} shows a representative example. 

\textbf{Direct-PMI.}
The model receives the pair $(x, y)$ and the definition of PMI, and returns a single scalar estimate $\widehat{\text{PMI}}(x, y)$. This tests whether an LLM can perform the full estimation task without decomposition. No conditional or marginal is elicited separately.
 
\textbf{Direct-Split.}
This method splits PMI into the two terms of Equation \ref{eq:pmi} and elicits each directly (Section \ref{sec:pmi_decomposition}). In the conditional prompt, the model sees an input $x$ and a candidate label $y$ and is asked to return an estimate of how likely this label is for this input, $P(y \mid x)$. The marginal term is estimated using the grounded base-rate prompt described in Section~\ref{sec:pmi_decomposition}. The final estimate combines the responses to these two prompts: 
$
\widehat{\text{PMI}}_{\text{Direct-Split}}(x, y) = \log \hat{P}(y \mid x) - \log \hat{P}(y).
$ 
 
\textbf{InfoNCE.}
Rather than asking for an open-ended probability, this method frames the conditional as a contrastive task (Section \ref{sec:contrastive}). The model sees an input $x$ and a candidate set $S = \{y_1, \ldots, y_K\}$ containing the true label and $K-1$ distractors sampled from the marginal. It assigns a probability to each candidate, and the probability on the true label serves as the score. Because the candidate set is closed, all probability mass is forced onto the listed options. This method estimates only the conditional term (and no marginal term):
$\widehat{\text{PMI}}_{\text{InfoNCE}}(x, y) = \log \hat{P}(y \mid x, S).
$
 
\textbf{MarginalNCE.}
Same contrastive conditional prompt as InfoNCE, but we subtract the shared grounded marginal described above:
$\widehat{\text{PMI}}_{\text{MarginalNCE}}(x, y) = \log \hat{P}(y \mid x, S) - \log \hat{P}(y).
$
This tests whether adding marginal correction to a contrastive conditional improves estimation, particularly on marginal-dominated datasets.
 
\textbf{PromptNCE.}
\textsc{PromptNCE} uses the same grounded marginal as Direct-Split and MarginalNCE, but modifies the contrastive conditional by adding an \textsc{other} category, so the model can place mass on unlisted labels rather than being forced to distribute all probability across a closed set. The final estimate is:
$
    \widehat{\text{PMI}}_{\text{PromptNCE}}(x, y) = \log \hat{P}(y \mid x, S \cup \{\textsc{other}\}) - \log \hat{P}(y; \text{grounded}).
$

\section{Benchmark}
\label{sec:benchmark}

We construct a benchmark from three publicly available datasets for evaluating
zero-shot PMI estimation from LLMs. Each instance is a text pair $(x,y)$ with
ground-truth PMI derived from human annotations (Table~\ref{tab:benchmark}).
The datasets differ in domain, label space size, and the relative importance of
the conditional and marginal terms of PMI. We evaluate with Spearman rank correlation, $\rho$, as described in Section~\ref{sec:challenge}.

\subsection{Datasets}
\label{sec:datasets}

Each dataset consists of input-label pairs $(x, y)$. Ground-truth PMI is computed as $\text{PMI}(x, y) = \log P(y \mid x) - \log P(y)$, with both terms derived from human annotations.

\textbf{Words.} The University of South Florida Free Association Norms \citep{nelson2004university}: participants see a cue word $x$ and produce the first word $y$ that comes to mind. $P(y \mid x)$ is the fraction of participants who produced $y$ given $x$; $P(y)$ is the overall frequency of $y$ across all cues.

\textbf{ChaosNLI.} Re-annotations of SNLI \citep{bowman2015large} premise-hypothesis pairs \citep{nie2020what}, each labeled by 100 independent annotators as entailment, neutral, or contradiction. Here $x$ is a premise-hypothesis pair and $y$ is an NLI label. $P(y \mid x)$ is the annotator vote share; $P(y)$ is the label frequency across the dataset.

\textbf{GoEmotions.} Reddit comments annotated with emotion labels by 3-5 raters \citep{demszky2020goemotions}. Here $x$ is a comment and $y$ is one of 28 emotion labels. $P(y \mid x)$ is the fraction of raters who selected $y$ for $x$; $P(y)$ is the overall frequency of $y$. Because raters may assign multiple emotions, this dataset differs from standard single-label classification.

\label{sec:variance_decomp}

The last three columns of Table~\ref{tab:benchmark} decompose PMI variance into
its conditional and marginal terms. The ratio $R = \text{Var}[\log P(y)] /
\text{Var}[\log P(y \mid x)]$ characterizes each dataset, and values greater
than 1 indicate a marginal-dominated dataset where PMI rankings are driven by
label base rates, $P(y)$. Words is strongly marginal-dominated ($R = 3.9$),
GoEmotions moderately so ($2.7$), and ChaosNLI is conditional-dominated
($0.07$). Ranking pairs on the empirical $\log P(y)$ alone, using ground-truth
label frequencies unavailable to our models, gives $\rho_{\mathrm{marg}} = 0.86$
on Words and $0.77$ on GoEmotions but $-0.03$ on ChaosNLI. On
marginal-dominated datasets, errors in $\hat{P}(y)$ propagate directly into PMI
rankings.

\begin{table}[t]
\centering
\footnotesize
\setlength{\tabcolsep}{4pt}
\begin{tabular}{lccccc}
\toprule
Dataset & \textsc{PromptNCE} & InfoNCE & Direct-Split & GEM 8B & GEM 70B \\
\midrule
Words
& \textbf{0.45}$\pm$0.04
& 0.21$\pm$0.04
& 0.35$\pm$0.04
& 0.24$\pm$0.05
& 0.43$\pm$0.04 \\

GoEmotions
& \textbf{0.27}$\pm$0.04
& 0.22$\pm$0.04
& 0.23$\pm$0.04
& 0.16$\pm$0.05
& 0.26$\pm$0.04 \\

ChaosNLI
& \textbf{0.80}$\pm$0.02
& 0.80$\pm$0.02
& 0.72$\pm$0.02
& 0.15$\pm$0.05
& 0.55$\pm$0.03 \\
\bottomrule
\end{tabular}
\caption{Spearman $\rho$ between the estimated conditional and human $P(y \mid
x)$, with no marginal term involved. $K=5$ except ChaosNLI ($K=3$). GEM is a
token-logprob baseline (Llama-3.1) and uses no candidate set. Bold marks the
best method per dataset. $n = 500$ pairs per dataset, Claude Sonnet 4. Values
are $\rho \pm$ standard error of the mean.}
\label{tab:diagnostic_ranking}
\end{table}

\section{Results}
\label{sec:results}

We evaluate on 500 pairs per dataset, disjoint from the 200 used during prompt development. All methods were run with GPT-5.2 \citep{openai2025gpt52} and Claude Sonnet 4 \citep{anthropic2025claude4}, specifically \texttt{gpt-5.2} and \texttt{claude-sonnet-4-20250514}. Claude Sonnet 4 consistently outperformed GPT-5.2 across methods and datasets. We report Claude Sonnet 4 results in this section. Full results for both models appear in the appendix. When elicited probabilities are reported as exactly zero, we clamp those probabilities to $10^{-6}$ before taking logarithms. 

\subsection{Conditional Probability Estimation}
\label{sec:cond_results}
Table \ref{tab:diagnostic_ranking} reports the conditional term alone,
scored against human $P(y \mid x)$ with no marginal involved.
\textsc{PromptNCE} gives the best conditional estimate on all three datasets, at
$\rho = 0.45$ on Words, $0.27$ on GoEmotions, and $0.80$ on ChaosNLI. The
closed-set contrastive prompt (InfoNCE) reaches $0.21$, $0.22$, and $0.80$, and
eliciting the conditional directly (Direct-Split) reaches $0.35$, $0.23$, and
$0.72$. The advantage of \textsc{PromptNCE} over the closed-set prompt is largest on
Words ($+0.24$), smaller on GoEmotions ($+0.05$), and disappears on ChaosNLI,
where the two methods tie at $\rho = 0.80$. GEM 70B is competitive on Words and GoEmotions, at $0.43$ and $0.26$,
and falls to $0.55$ on ChaosNLI. The smaller GEM 8B trails everywhere, at
$0.24$, $0.16$, and $0.15$. GEM reads probabilities off the decoder, so it needs
white-box access to token logprobs, while the prompted methods need only
black-box access to the model.

The datasets differ in how much of the label space $L$ the candidate set $K$ covers. At $K=5$, the candidates span $0.05\%$ of the $10{,}469$ responses in
Words and $18\%$ of the $28$ emotions in GoEmotions.  ChaosNLI has three labels, so its candidate set is the entire label space and \textsc{other} has no mass
left to absorb. The size of \textsc{PromptNCE}'s advantage follows this ordering,
from $+0.24$ on Words to $+0.05$ on GoEmotions to zero on ChaosNLI.

\subsection{End-to-End PMI Estimation}
\begin{figure}
    \centering
    \includegraphics[width=\linewidth]{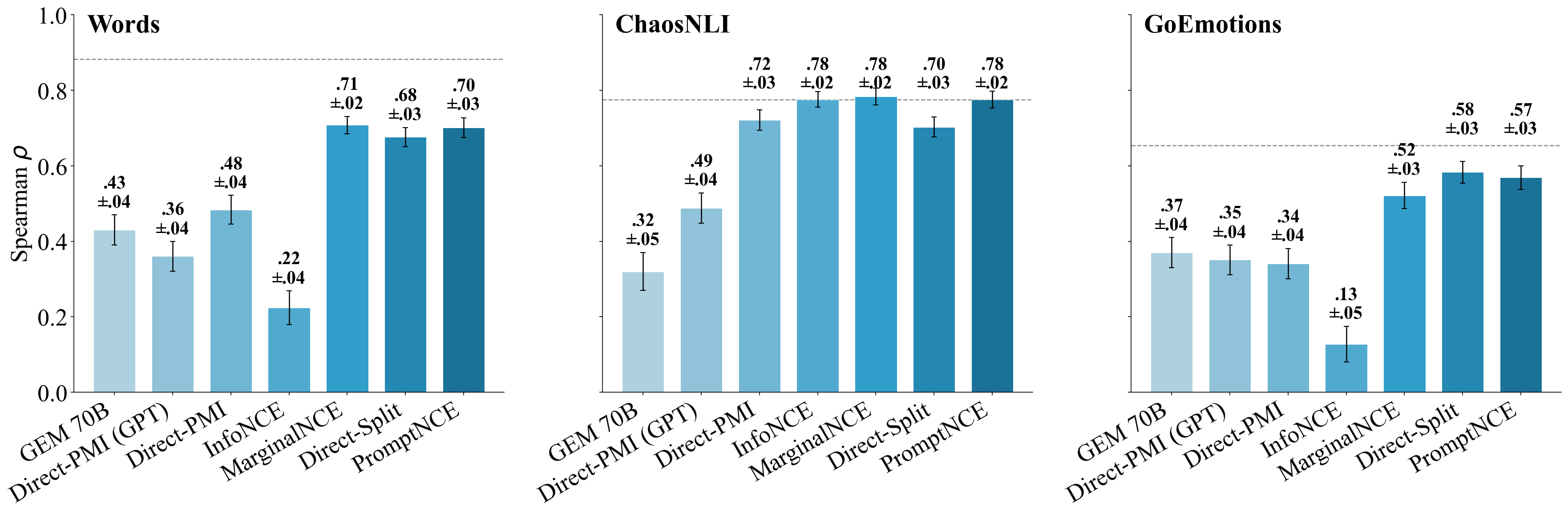}
    \caption{Spearman $\rho$ between estimated and true PMI, $K=5$ except ChaosNLI ($K=3$). All methods are Claude Sonnet 4 unless otherwise noted in the label. The dashed line shows \textsc{PromptNCE} using the empirical label marginal. Error bars are standard error of the mean.}
    \label{fig:main_results_bar}
\end{figure}

Figure \ref{fig:main_results_bar} reports Spearman $\rho$ between estimated and true PMI rankings across all methods and datasets. The best-performing methods vary across datasets: MarginalNCE reaches
$\rho=0.71$ on Words, Direct-Split reaches $0.58$ on GoEmotions, and
MarginalNCE, \textsc{PromptNCE}, and InfoNCE tie at $0.78$ on ChaosNLI. Direct-Split, MarginalNCE, and \textsc{PromptNCE} decompose PMI into
conditional and base-rate terms and share the same grounded marginal, differing
only in their conditional estimates. On Words and GoEmotions, their SEM
intervals overlap, at $0.68$--$0.71$ and $0.52$--$0.58$, so we cannot
distinguish their performance. On ChaosNLI, however, the two contrastive
conditionals, MarginalNCE and \textsc{PromptNCE}, reach $0.78$, with intervals that
do not overlap Direct-Split at $0.70$. This suggests that contrastive conditional estimation
helps when the conditional drives the ranking. \textsc{PromptNCE}'s conditional
advantage from Section~\ref{sec:cond_results} does not otherwise carry over to
full PMI. Methods without the grounded marginal fall behind, with InfoNCE
at $0.22$ on Words and $0.13$ on GoEmotions. Direct-PMI, which estimates PMI
without decomposition, is consistently weaker, though it reaches $\rho=0.72$
on ChaosNLI.

To investigate this further, we swept the candidate-set size $K$ from 5 to 20 on
GoEmotions, the only dataset where $K/L$ moves appreciably (Appendix
Table~\ref{tab:ksweep}). The elicited \textsc{other} mass falls from $0.34$ to
$0.11$, and \textsc{PromptNCE} declines from $\rho = 0.56$ to $0.47$, meeting
MarginalNCE within one SEM from $K = 10$ onward.

\subsection{Role of Marginal Estimation}
The variance decomposition predicts where marginal estimation will matter. We first verify this by examining the effect of adding a marginal term. InfoNCE, which uses only the conditional term, performs poorly on marginal-dominated datasets, reaching $\rho$ = 0.22 on Words and 0.13 on GoEmotions. Adding the grounded marginal term to InfoNCE gives MarginalNCE, which raises these to 0.71 and 0.52. On conditional-dominated ChaosNLI, the marginal correction has almost no effect, and both variants score $\rho$ = 0.78. This matches the variance decomposition. When $\text{Var}[\log P(y)]$ dominates, you need a good marginal, and when it does not, the marginal is noise.

To isolate the contribution of marginal estimation error, we replace the estimated marginal with the empirical dataset marginal, keeping conditional estimates unchanged. On Words, \textsc{PromptNCE} rises from 0.70 to 0.88 and Direct-Split from 0.68 to 0.89. On GoEmotions the gains are smaller but consistent, with \textsc{PromptNCE} rising from 0.57 to 0.65 and Direct-Split from 0.58 to 0.66. On ChaosNLI, the swap has essentially no effect. These results suggest that on marginal-dominated datasets, the gap between current performance and an achievable ceiling is largely attributable to marginal estimation error.

\subsection{Diagnosing Estimation Errors}
This section explores where errors come from, ranking or calibration, and if they reflect noise or systematic limits. We recalibrate each probability component using isotonic regression, which
preserves its rank order but can change the ranking of their difference. Gains are dataset-dependent and on GoEmotions they are large (Appendix Table \ref{tab:app_recalibration}). Because this requires a development set with ground-truth PMI,
we use recalibration only as a diagnostic. It is not viable under the low-data requirements of this challenge.

Ranking errors within the individual probability components, which recalibration cannot fix, are systematic, not stochastic. We ran the Direct-Split conditional prompt 10 times on 50 word-association pairs with caching disabled. The model's rankings are highly self-consistent across runs (mean pairwise $\rho$ = 0.86) but each run correlates with ground truth at only $\rho \approx 0.43$ (Appendix Table \ref{tab:app_stability}). Few-shot prompting does not close this gap either: 3-shot and 8-shot variants of the conditional prompt produce no consistent improvement over zero-shot across datasets or models (Appendix Table \ref{tab:app_prompt_interventions}).

\subsection{Case Study: Scoring Student Knowledge Summaries}
\label{sec:case_study}

This work evaluates zero-shot MI estimation on datasets with ground-truth PMI. But the methods would be helpful in settings where such ground truth does not exist. We present a case study where we use the contrastive methods to score summaries of student understanding in computer science education. LLMs can infer what students understand and struggle with from their work and express it in text. But how do we know whether these representations are accurate? Scoring them is hard for three reasons. Student programs are open-ended, so there is no fixed set of outcomes over which to predict what a
student writes next. Data on buggy student code is scarce and often not publicly available, making it difficult to train a generalizable critic. Moreover, we want an evaluation that works zero-shot, so it is accessible to teachers without model training.

A good summary should remain informative enough to identify the student it describes. We operationalize this as a contrastive task. Given a pool of code attempts, can the summary pick out the attempt belonging to the student it was written about? We score two
variants of this task. \emph{Compression} draws the candidate attempts from the same window the summary was written from, testing whether the summary retains enough student-specific detail to point back at its own source. \emph{Future prediction} draws them from a later attempt the summary was never shown, testing whether it captures something stable enough about the student to anticipate what she writes next. We test this on 100 students from a large online introductory programming course. An expert teacher wrote a short summary of each student based on four consecutive code snapshots. These expert summaries serve as a reference for an informative summary. A reliable scorer should assign higher scores to expert summaries than to uninformative ones. We therefore begin with this test, leaving the task of distinguishing among LLM-generated summaries for future work. For each student we score two summaries, the expert summary and a generic summary that could apply to any student. We score them with \textsc{PromptNCE}, InfoNCE, and GEM 8B.

\textsc{PromptNCE} reliably prefers the expert summary, scoring it above the generic summary on 88.0\% of compression comparisons and 73.5\% of future prediction comparisons, against 80.7\% and 65.0\% for InfoNCE (Table \ref{tab:case_study_gem}). GEM lands at 0.273 and 0.184, below the 0.50 chance rate, so the token-logprob score systematically prefers the generic summary to the expert one. We hypothesize that \textsc{PromptNCE}'s advantage here comes from the long-tailed nature of student code. The space of plausible future attempts is far larger than what can be captured in a small set of sampled negatives. A closed candidate set is therefore a poor approximation of the true output space. The explicit \textsc{other} category lets the model place most probability mass outside the listed candidates, producing conditional estimates that better reflect the open-ended nature of the task. This is the $K/L \approx 0$ regime of Section \ref{sec:cond_results}, where \textsc{PromptNCE}'s advantage is largest. This case study illustrates how zero-shot PMI estimation can provide an information-theoretic scoring function in settings where labeled data is too scarce to train a critic. We present it as a proof of concept rather than a validated evaluation method.

\begin{table}[t]
\centering
\small
\setlength{\tabcolsep}{6pt}
\begin{tabular}{lcc}
\toprule
Method & Compression & Future prediction \\
\midrule
\textsc{PromptNCE}          & $\mathbf{0.880}\pm0.019$ & $\mathbf{0.735}\pm0.026$ \\
InfoNCE                     & $0.807\pm0.024$ & $0.650\pm0.030$ \\
GEM 8B          & $0.273\pm0.035$ & $0.184\pm0.029$ \\
\bottomrule
\end{tabular}
\caption{Fraction of comparisons in which the expert summary scores above the generic summary (chance = 0.50). Each of 100 students contributes 4 comparisons for compression and up to 9 for future prediction, giving $n = 400$ and $n = 721$ respectively. Values are the fraction $\pm$ standard error of the mean, resampling at the student level.}
\label{tab:case_study_gem}
\end{table}

\section{Discussion and Conclusion}

\textbf{\textsc{PromptNCE} as a general tool for conditional probability
elicitation.}
The \textsc{PromptNCE} construction applies beyond PMI to any setting that
requires eliciting $P(y \mid x)$ from a prompted language model. Our results
suggest this is most useful when the candidate set covers only a small fraction
of the possible output space. \textsc{PromptNCE}'s advantage is largest on
Words, smaller on GoEmotions, and disappears on ChaosNLI when the candidates
span the full label space. Our case study provides a more open-ended example:
the space of possible student programs is effectively unbounded, and
\textsc{PromptNCE} outperforms closed-set InfoNCE and the GEM baseline \citep{xu2025gem} at scoring both compression and future prediction. Together, these results suggest that the \textsc{other} construction may be
useful for conditional probability elicitation when the output space is too
large or open-ended to enumerate.

\textbf{Limitations.} Our theoretical derivations assume a Bayes-optimal
classifier. Our experiments measure the gap between this ideal and actual LLM
behavior. We evaluate on three English-language datasets and two commercial
models, so conclusions about other domains, languages, and model families
require further validation. Our case study is limited to 100
students from a single course and cannot be released due to privacy
constraints.

\textsc{PromptNCE} also has several structural requirements. It
requires a candidate label set and a small number of unlabeled input-output
pairs. Because these examples come from the target dataset, the full estimator
is zero-training and low-data rather than strictly zero-shot; only its
conditional term is zero-shot. Its advantage also depends on the candidate set
covering a small fraction of the label space. As $K/L \to 1$, \textsc{other}
absorbs little mass and \textsc{PromptNCE} approaches the closed-set estimator.
Likewise, when PMI rankings are dominated by the marginal, improved conditional
estimation has little effect.

The \textsc{other} category may absorb probability assigned to
near-synonyms of $y$ \citep{holtzman2021surface}. This is correct for the exact
event $P(Y=y\mid X=x)$ our prompts request, but depends on the model respecting
that framing. Because PMI combines independently elicited conditional and
marginal terms, errors in either can distort the final ranking. Finally, we
cannot rule out benchmark contamination because model pretraining corpora are
not public, and our variance decomposition requires ground-truth PMI, making it
a retrospective diagnostic rather than a deployment-time tool.

\textbf{Conclusion.} We introduced the Zero-Shot LLM Mutual Information Estimation Challenge 
and showed that prompted language models can estimate pointwise mutual information without training data, achieving Spearman $\rho$ up to 0.78 against human-derived ground truth. Our key contribution is \textsc{PromptNCE},
which augments the contrastive candidate set with an explicit
\textsc{other} category so that the model can assign probability mass outside the closed candidate set. Empirically, \textsc{PromptNCE} gives the best conditional estimates on every
dataset we tested, with an advantage that tracks how little of the label space
the candidate set covers. A variance decomposition of dataset structure predicts when methods succeed and where they fail. Beyond PMI, the \textsc{other} construction offers a general and lightweight technique for improving conditional 
probability elicitation from prompted LLMs --- a finding we hope will be useful well beyond this benchmark.

\section*{Acknowledgments}
We would like to sincerely thank the Carina Foundation for making this research
possible.

\bibliography{colm2026_conference}
\bibliographystyle{colm2026_conference}

\appendix

\renewcommand{\thetable}{\Alph{section}\arabic{table}}
\renewcommand{\thefigure}{\Alph{section}\arabic{figure}}

\section{Disclosure on LLM Usage}
LLMs were used to improve and polish writing for clarity and conciseness, as well as to generate LaTeX code for tables. We also used LLMs to support our exploration and understanding of existing mutual information estimation methods, both in identifying relevant prior work and in clarifying aspects of the methods as we read the original papers. LLMs were additionally used to streamline parts of the coding process. The authors have thoroughly reviewed all LLM-generated content and take full responsibility for everything in this work.

\section{Additional results}
\setcounter{table}{0}
\setcounter{figure}{0}

\subsection{Full results across models}
\label{app:full_results}

Table~\ref{tab:app_full_results} reports Spearman $\rho$ for all methods on both GPT-5.2 and Claude Sonnet 4. Claude Sonnet 4 consistently outperforms GPT-5.2 across methods and datasets. The main paper results report Claude Sonnet 4.

\begin{table*}[t]
\centering
\tiny
\begin{tabular}{l cccccc}
\toprule
 & \multicolumn{2}{c}{\textbf{Words}} & \multicolumn{2}{c}{\textbf{ChaosNLI}} & \multicolumn{2}{c}{\textbf{GoEmotions}} \\
\cmidrule(lr){2-3} \cmidrule(lr){4-5} \cmidrule(lr){6-7}
Method & GPT-5.2 & Claude & GPT-5.2 & Claude & GPT-5.2 & Claude \\
\midrule
\multicolumn{7}{l}{\textit{Grounded marginal, $K=5$ (the configuration reported in Figure \ref{fig:main_results_bar})}} \\[2pt]
Direct-Split (grounded)
& 0.67$\pm$0.03 & 0.68$\pm$0.03
& 0.69$\pm$0.03 & 0.70$\pm$0.03
& 0.49$\pm$0.03 & 0.58$\pm$0.03 \\
MarginalNCE (grounded)
& 0.74$\pm$0.02 & 0.71$\pm$0.02
& 0.71$\pm$0.03 & 0.78$\pm$0.02
& 0.45$\pm$0.04 & 0.52$\pm$0.03 \\
PromptNCE (grounded)
& 0.74$\pm$0.02 & 0.70$\pm$0.03
& 0.73$\pm$0.03 & 0.78$\pm$0.02
& 0.48$\pm$0.04 & 0.57$\pm$0.03 \\
\midrule
\multicolumn{7}{l}{\textit{Empirical marginal (oracle upper bound)}, $K=25$} \\[2pt]
Direct-Split (emp.)          & 0.82$\pm$0.02 & 0.89$\pm$0.01 & 0.69$\pm$0.03 & 0.70$\pm$0.03 & 0.56$\pm$0.03 & 0.66$\pm$0.03 \\
InfoNCE (emp.)      & 0.87$\pm$0.01 & 0.89$\pm$0.01 & 0.70$\pm$0.03 & 0.73$\pm$0.03 & 0.53$\pm$0.03 & 0.61$\pm$0.03 \\
PromptNCE (emp.) & 0.85$\pm$0.02 & 0.85$\pm$0.02 & 0.73$\pm$0.03 & 0.81$\pm$0.02 & 0.49$\pm$0.04 & 0.55$\pm$0.04 \\
\midrule
\multicolumn{7}{l}{\textit{Zero-shot methods}, model-estimated marginal, $K=25$} \\[2pt]
InfoNCE              & 0.35$\pm$0.04 & 0.33$\pm$0.04 & 0.69$\pm$0.03 & 0.73$\pm$0.03 & 0.13$\pm$0.05 & 0.18$\pm$0.05 \\
MarginalNCE & 0.64$\pm$0.03 & 0.64$\pm$0.03 & 0.69$\pm$0.03 & 0.73$\pm$0.03 & 0.41$\pm$0.04 & 0.45$\pm$0.04 \\
PromptNCE        & \textbf{0.74}$\pm$0.02 & \textbf{0.69}$\pm$0.03 & \textbf{0.73}$\pm$0.03 & \textbf{0.82}$\pm$0.02 & 0.43$\pm$0.04 & \textbf{0.47}$\pm$0.04 \\
Direct-Split                  & 0.57$\pm$0.03 & 0.63$\pm$0.03 & 0.70$\pm$0.03 & 0.75$\pm$0.03 & \textbf{0.44}$\pm$0.04 & 0.40$\pm$0.04 \\
Direct-PMI           & 0.36$\pm$0.04 & 0.48$\pm$0.04 & 0.49$\pm$0.04 & 0.72$\pm$0.03 & 0.35$\pm$0.04 & 0.34$\pm$0.04 \\
\bottomrule
\end{tabular}
\caption{Spearman $\rho$ between estimated and true PMI rankings (500 held-out pairs per dataset). The first block gives every decomposed method the same grounded marginal at $K=5$ and reproduces Figure \ref{fig:main_results_bar}, whose Claude column is the main-paper result. The lower two blocks use $K=25$, and their zero-shot methods estimate the marginal from the model's own base-rate prompt rather than the grounded one, which is why the same method names carry different values across blocks. \textbf{Bold} indicates best zero-shot method per column within the model-estimated block. Error bars are bootstrap standard errors.}
\label{tab:app_full_results}
\end{table*}

\subsection{Diagnostic decomposition across models}
\label{app:diagnostic}

Tables \ref{tab:app_diagnostic_cond} and \ref{tab:app_diagnostic_marg} extend the main-paper diagnostic (Table \ref{tab:diagnostic_ranking}) to both models at $K=25$, where the main-paper table uses $K=5$. Only the contrastive conditionals move with $K$. Direct-Split elicits its conditional without a candidate set, so its values match the main-paper table exactly. The pattern is consistent across both settings. PromptNCE achieves the best conditional ranking on all datasets for both models, and its grounded marginal prompt outperforms the open-ended baseline. Calibration slopes (ideal value 1.0) vary across methods, and post-hoc recalibration recovers real gains on GoEmotions while leaving ranking errors untouched (Section \ref{app:recalibration}).

\begin{table}[t]
\centering
\small
\begin{tabular}{l cccc}
\toprule
 & \multicolumn{2}{c}{Rank $\rho$} & \multicolumn{2}{c}{Cal.\ slope} \\
\cmidrule(lr){2-3} \cmidrule(lr){4-5}
Method & GPT-5.2 & Claude & GPT-5.2 & Claude \\
\midrule
\multicolumn{5}{l}{\textit{Words}} \\[2pt]
Direct-Split          & 0.33$\pm$0.04 & 0.35$\pm$0.04 & 0.55$\pm$0.05 & 0.32$\pm$0.03 \\
InfoNCE      & 0.26$\pm$0.04 & 0.38$\pm$0.04 & 0.25$\pm$0.03 & 0.23$\pm$0.02 \\
PromptNCE & 0.38$\pm$0.04 & 0.50$\pm$0.04 & 0.40$\pm$0.05 & 0.78$\pm$0.08 \\
\midrule
\multicolumn{5}{l}{\textit{ChaosNLI}} \\[2pt]
Direct-Split          & 0.67$\pm$0.03 & 0.72$\pm$0.03 & 0.59$\pm$0.03 & 0.48$\pm$0.02 \\
InfoNCE      & 0.73$\pm$0.02 & 0.76$\pm$0.03 & 0.72$\pm$0.03 & 0.80$\pm$0.04 \\
PromptNCE & 0.75$\pm$0.02 & 0.83$\pm$0.02 & 0.75$\pm$0.03 & 0.92$\pm$0.09 \\
\midrule
\multicolumn{5}{l}{\textit{GoEmotions}} \\[2pt]
Direct-Split          & 0.30$\pm$0.04 & 0.23$\pm$0.04 & 0.65$\pm$0.09 & 0.36$\pm$0.06 \\
InfoNCE      & 0.34$\pm$0.04 & 0.31$\pm$0.04 & 0.96$\pm$0.11 & 0.73$\pm$0.10 \\
PromptNCE & 0.34$\pm$0.04 & 0.33$\pm$0.04 & 0.95$\pm$0.19 & 1.63$\pm$0.25 \\
\bottomrule
\end{tabular}
\caption{Diagnostic decomposition: conditional $\hat{P}(y \mid x)$. Spearman $\rho$ and calibration slope between estimated and true conditional probabilities (500 pairs per dataset, $K=25$). Error bars are standard error of the mean.}
\label{tab:app_diagnostic_cond}
\end{table}

\begin{table}[t]
\centering
\small
\begin{tabular}{l cccc}
\toprule
 & \multicolumn{2}{c}{Rank $\rho$} & \multicolumn{2}{c}{Cal.\ slope} \\
\cmidrule(lr){2-3} \cmidrule(lr){4-5}
Method & GPT-5.2 & Claude & GPT-5.2 & Claude \\
\midrule
\multicolumn{5}{l}{\textit{Words}} \\[2pt]
Direct-Split / InfoNCE & 0.63$\pm$0.03 & 0.63$\pm$0.03 & 0.53$\pm$0.04 & 0.63$\pm$0.05 \\
PromptNCE  & 0.81$\pm$0.02 & 0.73$\pm$0.02 & 0.65$\pm$0.03 & 0.57$\pm$0.03 \\
\midrule
\multicolumn{5}{l}{\textit{GoEmotions}} \\[2pt]
Direct-Split / InfoNCE & 0.42$\pm$0.20 & 0.44$\pm$0.19 & 0.35$\pm$0.13 & 0.35$\pm$0.10 \\
PromptNCE  & 0.73$\pm$0.13 & 0.74$\pm$0.12 & 1.03$\pm$0.09 & 0.93$\pm$0.13 \\
\bottomrule
\end{tabular}
\caption{Diagnostic decomposition: marginal $\hat{P}(y)$. Spearman $\rho$ and calibration slope between estimated and true marginal probabilities (500 pairs per dataset, $K=25$). Direct-Split and InfoNCE share the same marginal prompt, so their rows are merged. ChaosNLI is omitted because its 3-label marginal makes ranking ill-defined. Error bars are standard error of the mean.}
\label{tab:app_diagnostic_marg}
\end{table}

\subsection{Candidate-set size sweep}
\label{app:ksweep}

Table~\ref{tab:ksweep} holds the dataset fixed and sweeps the
candidate-set size $K$. GoEmotions is the only dataset where $K/L$ moves
appreciably: Words has $L = 10{,}469$, so $K/L \approx 0$ at every $K$, and
ChaosNLI is pinned at $K = L = 3$.

\begin{table}[t]
\centering
\begingroup
\footnotesize
\setlength{\tabcolsep}{5pt}
\begin{tabular}{lcccc}
\toprule
& $K=5$ & $K=10$ & $K=15$ & $K=20$ \\
\midrule
InfoNCE
& $0.15 \pm 0.04$ & $0.20 \pm 0.04$ & $0.17 \pm 0.04$ & $0.19 \pm 0.04$ \\
MarginalNCE
& $0.50 \pm 0.04$ & $\mathbf{0.51 \pm 0.04}$ & $\mathbf{0.49 \pm 0.04}$ & $\mathbf{0.47 \pm 0.04}$ \\
PromptNCE
& $\mathbf{0.56 \pm 0.03}$ & $\mathbf{0.52 \pm 0.04}$ & $\mathbf{0.50 \pm 0.04}$ & $\mathbf{0.47 \pm 0.04}$ \\
\midrule
$K/L$ coverage
& $0.179$ & $0.357$ & $0.536$ & $0.714$ \\
OTHER mass
& $0.340$ & $0.206$ & $0.140$ & $0.105$ \\
\bottomrule
\end{tabular}
\endgroup
\caption{Spearman $\rho$ against \emph{full} PMI on GoEmotions as candidate-set
size $K$ increases. The final row reports \textsc{PromptNCE}'s elicited
\textsc{other} mass. Bold marks the best method per column. Claude Sonnet 4.
Values are $\rho \pm$ standard error of the mean.}
\label{tab:ksweep}
\end{table}

\subsection{Prompting interventions across models}
\label{app:prompting}

Table~\ref{tab:app_prompt_interventions} shares a small prompting intervention analysis on 80 pairs from two datasets. The pattern holds for both models: no intervention produces a consistent improvement over the baseline, the variation across interventions is smaller than the variation across models, and distribution-aware prompting actively degrades performance.

\begin{table}[t]
\centering
\small
\begin{tabular}{l cc cc}
\toprule
& \multicolumn{2}{c}{\textbf{Words}} & \multicolumn{2}{c}{\textbf{ChaosNLI}} \\
\cmidrule(lr){2-3} \cmidrule(lr){4-5}
\textbf{Intervention} & GPT-5.2 & Claude & GPT-5.2 & Claude \\
\midrule
Baseline              & .39 {\scriptsize$\pm$.10} & .44 {\scriptsize$\pm$.09} & .77 {\scriptsize$\pm$.06} & .84 {\scriptsize$\pm$.03} \\
Generate-then-rank    & .40 {\scriptsize$\pm$.10} & .46 {\scriptsize$\pm$.09} & .82 {\scriptsize$\pm$.04} & .84 {\scriptsize$\pm$.05} \\
Log-scale prior       & .40 {\scriptsize$\pm$.10} & .46 {\scriptsize$\pm$.09} & .84 {\scriptsize$\pm$.04} & .81 {\scriptsize$\pm$.05} \\
3-shot                & .38 {\scriptsize$\pm$.10} & .50 {\scriptsize$\pm$.09} & .79 {\scriptsize$\pm$.05} & .86 {\scriptsize$\pm$.03} \\
8-shot                & .45 {\scriptsize$\pm$.09} & .37 {\scriptsize$\pm$.10} & .79 {\scriptsize$\pm$.05} & .80 {\scriptsize$\pm$.05} \\
Distribution-aware    & .30 {\scriptsize$\pm$.10} & .33 {\scriptsize$\pm$.10} & .59 {\scriptsize$\pm$.09} & .76 {\scriptsize$\pm$.06} \\
\bottomrule
\end{tabular}
\caption{Conditional ranking Spearman $\rho$ (Direct-Split) across prompting interventions and models (80 held-out pairs per dataset). No intervention produces a consistent improvement over the baseline. Distribution-aware prompting actively degrades performance. Error bars are bootstrapped standard error of the mean.}
\label{tab:app_prompt_interventions}
\end{table}

\subsection{Post-hoc recalibration}
\label{app:recalibration}

To assess whether calibration errors contribute meaningfully to PMI estimation quality, we apply isotonic regression to the conditional estimates, the marginal estimates, or both, fitting on a 200-pair development set and evaluating on 500 held-out pairs. Table~\ref{tab:app_recalibration} reports the results. Gains vary by dataset: small on ChaosNLI, moderate on Words, and substantial on GoEmotions, where isotonic regression on both components raises \textsc{PromptNCE} from $\rho$ = 0.47 to 0.67 and Direct-Split from 0.40 to 0.57. Calibration error therefore contributes meaningfully, most clearly where the conditional is weakest. Fitting the isotonic map requires ground-truth PMI on a development set, so these numbers bound what better calibration could achieve rather than describing a method available zero-shot. Ranking errors within an individual probability component are what recalibration cannot fix, since isotonic regression is monotone and preserves its rank order by construction. For comparison, we also report an oracle condition that replaces the LLM-estimated marginal with the true empirical marginal, which provides a ceiling on the gains achievable from better marginal estimation alone.

\begin{table*}[t]
\centering
\small
\setlength{\tabcolsep}{4pt}
\begin{tabular}{ll cc cc cc}
\toprule
 & & \multicolumn{2}{c}{Words} & \multicolumn{2}{c}{ChaosNLI} & \multicolumn{2}{c}{GoEmotions} \\
\cmidrule(lr){3-4} \cmidrule(lr){5-6} \cmidrule(lr){7-8}
Method & Recalibration & GPT-5.2 & Claude & GPT-5.2 & Claude & GPT-5.2 & Claude \\
\midrule
\multirow{4}{*}{Direct-Split}
 & Uncalibrated       & 0.57 & 0.63 & 0.70 & 0.75 & 0.44 & 0.40 \\
 & Isotonic (cond.)    & \textbf{0.65} & 0.64 & 0.70 & 0.72 & \textbf{0.58} & 0.56 \\
 & Isotonic (both)    & 0.65 & \textbf{0.65} & 0.69 & 0.72 & 0.53 & \textbf{0.57} \\
 & Empirical marginal & 0.82 & 0.89 & 0.69 & 0.70 & 0.56 & 0.66 \\
\midrule
\multirow{4}{*}{PromptNCE}
 & Uncalibrated       & 0.74 & 0.69 & 0.73 & 0.82 & 0.43 & 0.47 \\
 & Isotonic (cond.)    & \textbf{0.80} & \textbf{0.76} & 0.73 & 0.81 & 0.62 & 0.66 \\
 & Isotonic (both)    & 0.79 & 0.76 & 0.73 & 0.81 & \textbf{0.67} & \textbf{0.67} \\
 & Empirical marginal & 0.85 & 0.85 & 0.73 & 0.81 & 0.49 & 0.55 \\
\bottomrule
\end{tabular}
\caption{Effect of post-hoc recalibration on PMI estimation (Spearman $\rho$). Isotonic regression is fit on a 200-pair development set and evaluated on 500 held-out pairs, $K=25$. \textbf{Bold} indicates best recalibrated result per method and dataset. The empirical marginal row replaces the LLM's $\hat{P}(y)$ with the true marginal, providing a ceiling on marginal-estimation gains.}
\label{tab:app_recalibration}
\end{table*}

\subsection{Estimation stability}
\label{app:stability}

To determine whether conditional ranking errors are stochastic (reducible by averaging) or systematic (reflecting limits in the model's knowledge), we ran the Direct-Split conditional prompt 10 times on 50 word-association pairs with API caching disabled. Table~\ref{tab:app_stability} reports the results. The model's rankings are highly self-consistent across runs (mean pairwise $\rho = 0.86$) but each run correlates with ground truth at only $\rho \approx 0.43$. Averaging estimates across all 10 runs does not improve agreement with ground truth ($\rho = 0.44$), confirming that the errors are systematic rather than stochastic.

\begin{table}[t]
\centering
\small
\begin{tabular}{lc}
\toprule
Metric & Value \\
\midrule
\multicolumn{2}{l}{\textit{Per-pair variability (CV across 10 runs)}} \\
\quad Median CV & 0.35 \\
\quad Mean CV & 0.40 \\
\quad IQR & [0.30, 0.48] \\
\midrule
\multicolumn{2}{l}{\textit{Inter-run rank agreement (Spearman $\rho$)}} \\
\quad Mean & 0.86 \\
\quad Range & [0.76, 0.95] \\
\midrule
\multicolumn{2}{l}{\textit{Agreement with ground truth (Spearman $\rho$)}} \\
\quad Per-run mean & 0.43 \\
\quad Per-run range & [0.38, 0.48] \\
\quad Averaged estimate & 0.44 \\
\bottomrule
\end{tabular}
\caption{Stability of GPT-5.2's $\hat{P}(y \mid x)$ estimates on 50 word-association pairs, each prompted 10 times. The model is highly self-consistent across runs (inter-run $\rho = 0.86$) but consistently misranked relative to ground truth ($\rho \approx 0.43$). Averaging across runs does not improve agreement, confirming that ranking errors are systematic.}
\label{tab:app_stability}
\end{table}

\section{Example Prompts}
\label{app:prompts}

We include representative prompts for each method, shown across different datasets so readers can see how the task framing adapts. All prompts for all methods and datasets are available at \url{https://juliettewoodrow.github.io/promptnce/}. MarginalNCE is not shown separately because it combines the InfoNCE conditional prompt (Section \ref{app:infonce}) with the Direct-Split marginal prompt (Section \ref{app:decomp_marginal}).
 
\subsection{Direct-PMI (GoEmotions)}
\label{app:direct}
 
\begin{promptbox}
You are simulating a large-scale human annotation study of Reddit comments.
Each rater reads ONE comment and selects ALL emotions that apply from a fixed set.
Raters may select multiple emotions.
 
Define an event T as: a randomly chosen rater selects the TARGET emotion for this comment.
 
Pointwise mutual information (PMI) in nats is: PMI = ln(P(T|comment) / P(T)).
 
COMMENT:
\{comment\}
 
TARGET EMOTION: \{target\}
 
Return ONLY JSON: \{"PMI\_LN": <number>, "notes": "<short>"\}.
\end{promptbox}
 
\subsection{Direct-Split Conditional Prompt (ChaosNLI)}
\label{app:decomp_conditional}
\begin{promptbox}
In a large NLI annotation study, 100 raters each read a premise-hypothesis pair and choose ONE label: entailment, neutral, or contradiction. Estimate p\_apply = P(a randomly chosen rater selects TARGET label for this pair). \\
\\
PREMISE: \{premise\} \\
HYPOTHESIS: \{hypothesis\} \\
TARGET LABEL: \{target\}\\
\\
Return ONLY JSON: \{"p\_apply": <number>, "notes": "<short>"\}. \\
p\_apply must be in (0,1].
\end{promptbox}
 
\subsection{Direct-Split Marginal Prompt (Words)}
\label{app:decomp_marginal}
 
\begin{promptbox}
In a free-association task, estimate p\_base = P(TARGET is first response) over random cues.

TARGET: \{target\} \\
 
Return ONLY JSON: \{"p\_base": <number>, "notes": "<short>"\}. \\
p\_base must be in (0,1].
\end{promptbox}
 
\subsection{InfoNCE Conditional Prompt (GoEmotions)}
\label{app:infonce}
 
\begin{promptbox}
You are simulating a human emotion-annotation task on Reddit comments.
A rater reads ONE comment and chooses exactly ONE emotion label from the list provided. Treat the list as a CLOSED set for this question. \\
 \\
COMMENT:
\{comment\} \\
 
Emotion options (closed set): \\
- \{option\_1\} \\
- \{option\_2\} \\
- ... \\
- \{option\_K\} \\
 
Return ONLY JSON mapping each option to probability. Probabilities must sum to 1.
\end{promptbox}

\subsection{PromptNCE Conditional Prompt (Words)}

\begin{promptbox}  
You are simulating a large-sample human free-association study. A participant sees ONE cue word and says the FIRST response word that comes to mind.

CUE: \{cue\_word\}

Candidate response words (partial list) \\
\{option\_1\}, \{option\_2\}, ..., \{option\_K\} \\

TASK:
Return a probability mass function (PMF) estimating how likely each candidate is to be the participant's first response. Include a special key "OTHER" for the probability that the response is some word NOT in the candidate list.

Calibration guidance: \\
- The candidate list is partial (not exhaustive). Many response words
  are not listed. \\
- Strong associates should get high probability; weak ones near zero. \\
- Use OTHER for remaining probability mass (unlisted responses). \\

OUTPUT REQUIREMENTS: \\
- Output ONLY the JSON object. \\
- Include EVERY candidate word plus OTHER. \\
- Probabilities must be numeric and sum to 1. \\
\end{promptbox}

\subsection{PromptNCE Marginal Prompt (Words)}

\begin{promptbox}
You are simulating a large-sample human free-association study.
A participant sees a random cue word and says the FIRST response word. \\

We want the BASE RATE probability that a specific target word is said
as a first response, averaged across ALL possible cue words. \\

Here are some example target words used in this study (no frequencies given): \{example\_label\_1\}, \{example\_label\_2\}, ... \\

Here are a few example cue words and their common responses
(for grounding only; do NOT assume frequencies):

Example 1: CUE = \{example\_cue\_1\} \\
  Common responses: \{responses\_1\} \\
Example 2: CUE = \{example\_cue\_2\} \\
  Common responses: \{responses\_2\} \\

TARGET WORD: \{target\} \\

TASK: \\
Estimate p\_base = P(TARGET WORD is said as first response) across random cue words. \\

Important notes: \\
- Very common response words (e.g. WATER, LOVE) have p\_base ~ 0.001-0.01. \\
- Most words have p\_base ~ 0.00001-0.0001. \\
- Rare words can be as low as 0.000001. \\
- p\_base must be in (0,1]; do NOT use 0. \\

Return ONLY JSON: {"p\_base": <number>}
\end{promptbox}

\end{document}